# Multimodal spatiotemporal graph neural networks for improved prediction of 30-day all-cause hospital readmission

Siyi Tang[1,*], Amara Tariq[2,*], Jared Dunnmon[3], Umesh Sharma[4], Praneetha Elugunti[5], Daniel Rubin[3,6], Bhavik N. Patel[2,5,†], Imon Banerjee[2,†,‡]

[1]Department of Electrical Engineering, Stanford University, CA, USA
[2]Department of Radiology, Mayo Clinic, Phoenix, AZ, USA
[3]Department of Biomedical Data Science, Stanford University, CA, USA
[4]Department of Medicine, Division of Hospital Internal Medicine, Mayo Clinic, AZ, USA
[5]Department of Administration, Mayo Clinic, AZ, USA
[6]Department of Radiology, Stanford University, CA, USA

[*]Contributed equally as first authors
[†]Contributed equally as senior authors
[‡]Corresponding author

**Address correspondence to:**
Imon Banerjee
Banerjee.Imon@mayo.edu

**Running title:**
Multimodal graph model for readmission prediction

# Abstract

Measures to predict 30-day readmission are considered an important quality factor for hospitals as accurate predictions can reduce the overall cost of care by identifying high risk patients before they are discharged. While recent deep learning-based studies have shown promising empirical results on readmission prediction, several limitations exist that may hinder widespread clinical utility, such as: (a) only patients with certain conditions are considered, (b) existing approaches do not leverage data temporality, (c) individual admissions are assumed independent of each other, which is unrealistic, (d) prior studies are usually limited to single source of data, or (e) existing methods are limited to single center data. To address these limitations, we propose a multimodal, modality-agnostic spatiotemporal graph neural network (MM-STGNN) for prediction of 30-day all-cause hospital readmission that fuses multimodal in-patient longitudinal data. By training and evaluating our methods using longitudinal chest radiographs and electronic health records from two independent centers, we demonstrate that MM-STGNN achieves an area under the receiver operating characteristic curve (AUROC) of 0.79 on both primary and external datasets. Furthermore, MM-STGNN significantly outperforms the current clinical reference standard, LACE+ score (AUROC=0.61), on the primary dataset. For subset populations of patients with heart and vascular disease, our model also outperforms baselines, such as gradient-boosting model, non-temporal graph model, and Long Short-Term Memory network, on predicting 30-day readmission (e.g., 3.7 point improvement in AUROC in patients with heart disease). Lastly, qualitative model interpretability analysis indicates that while patients' primary diagnoses were not explicitly used to train the model, node features crucial for model prediction directly reflect patients' primary diagnoses. Importantly, our MM-STGNN is agnostic to node feature modalities and could be utilized to integrate multimodal data for triaging patients in various downstream resource allocation tasks.

# Introduction

Hospital readmission rates are used as one of the primary quality measures in the United States healthcare system. Unplanned hospital readmissions within a short period of discharge (e.g., 30 days) impose a substantial burden on patients and healthcare systems[1]. An example of such unplanned readmissions can be readmission to the hospital for a surgical wound infection that occurred after the initial hospital stay. In the United States, approximately 1.8 million Medicare beneficiaries aged above 65 years old experience a readmission, which costs Medicare $26 billion every year[2]. However, it is estimated that $17 billion of the readmission costs comes from potentially preventable readmissions[2]. Therefore, reducing avoidable hospital readmissions has become a major goal for healthcare systems.

Accurate hospital readmission prediction could identify high risk patients and potentially reduce avoidable readmission costs for healthcare systems. Moreover, accurately predicting hospital readmissions could improve outcomes by identifying patients who are at high risk of poor post-discharge outcomes and allowing healthcare workers to either provide opportunities for closer follow-up after discharge or early intervention. Several rule-based or linear risk score models have been adopted clinically to predict hospital readmissions, such as the LACE+ index[3] and the HOSPITAL[4] score. However, these risk scores are calculated based on a limited number of pre-defined clinical variables and have been shown to have only poor-to-moderate discriminative power in identifying patients at high-risk of being readmitted, particularly for young patients with rare diseases[5,6].

In contrast, deep learning (DL) models are capable of learning from large amounts of minimally processed data and can ingest data from multiple modalities, such as patient clinical and imaging data, to potentially improve the prediction accuracy. In recent years, DL techniques have been widely applied for predicting 30-day hospital readmissions[7–18], with area under the receiver operating characteristics curve (AUROC) ranging from 0.64 to 0.90 across a wide variety of populations.

Despite promising empirical results, several major limitations exist in prior studies. First, many are limited to patients with a certain subset of conditions[7,11,12,14,17] (e.g., congestive heart failure), which considers more homogeneous patient populations but restricts the applicability of the prediction models to only a subpopulation. Second, while temporal relationships in patient data are crucial for predicting outcomes after hospital discharge, only a few studies have leveraged longitudinal data[7,14,17]. Third, most of the prior studies treat individual patients or hospitalizations as independent data points, which ignores the relationship between patients or their hospitalizations. To our knowledge, only one study has leveraged graph neural networks (GNNs) to model the topological structure between patients for hospital readmission prediction[18]; however, it only utilizes clinical notes and does not consider the temporality of hospital admission data. Fourth, while patient data often involve multiple modalities, such as imaging and clinical data, only a few studies have included multimodal data for hospital readmission prediction[15,16]. Lastly, few studies have validated the proposed methods on multicenter data[13].

In this study, we address the aforementioned limitations by proposing a multimodal spatiotemporal graph neural network (MM-STGNN) for 30-day all-cause hospital readmission prediction from patient longitudinal imaging and electronic health records (EHR) data. Our main contributions are as follows: (a) we propose a novel representation of topological relationship between hospital admissions using a graph, where each node is one hospital admission and the edges represent the similarity between admissions based on patient characteristics; (b) we design a multimodal, modality-agnostic spatiotemporal graph neural network (MM-STGNN) that can ingest multiple modalities of in-patient longitudinal data; (c) finally, we perform model interpretability analysis to explain which features are important in the model's prediction. By training and evaluating our methods on two large datasets from independent medical centers with longitudinal imaging and EHR data, we show that our MM-STGNN achieves an AUROC of 0.79 on both primary and external datasets. Importantly, MM-STGNN significantly outperforms the current clinical reference standard, the LACE+ score (AUROC=0.61), providing 15.8 points improvement in AUROC on the primary dataset. Compared to baselines such as gradient-boosting model, non-temporal graph model, and recurrent neural network, our MM-STGNN improves the prediction performance of disease-specific subpopulations, including patients with heart and vascular diseases (e.g., 3.7 points improvement in AUROC over baselines on patients with heart diseases). Lastly, while patients' primary diagnoses were not explicitly used to train MM-STGNN, qualitative model interpretability analysis suggests that the most crucial node features for model prediction reflect the patient primary diagnoses and corresponding treatments, which could facilitate clinicians' understanding of model decisions.

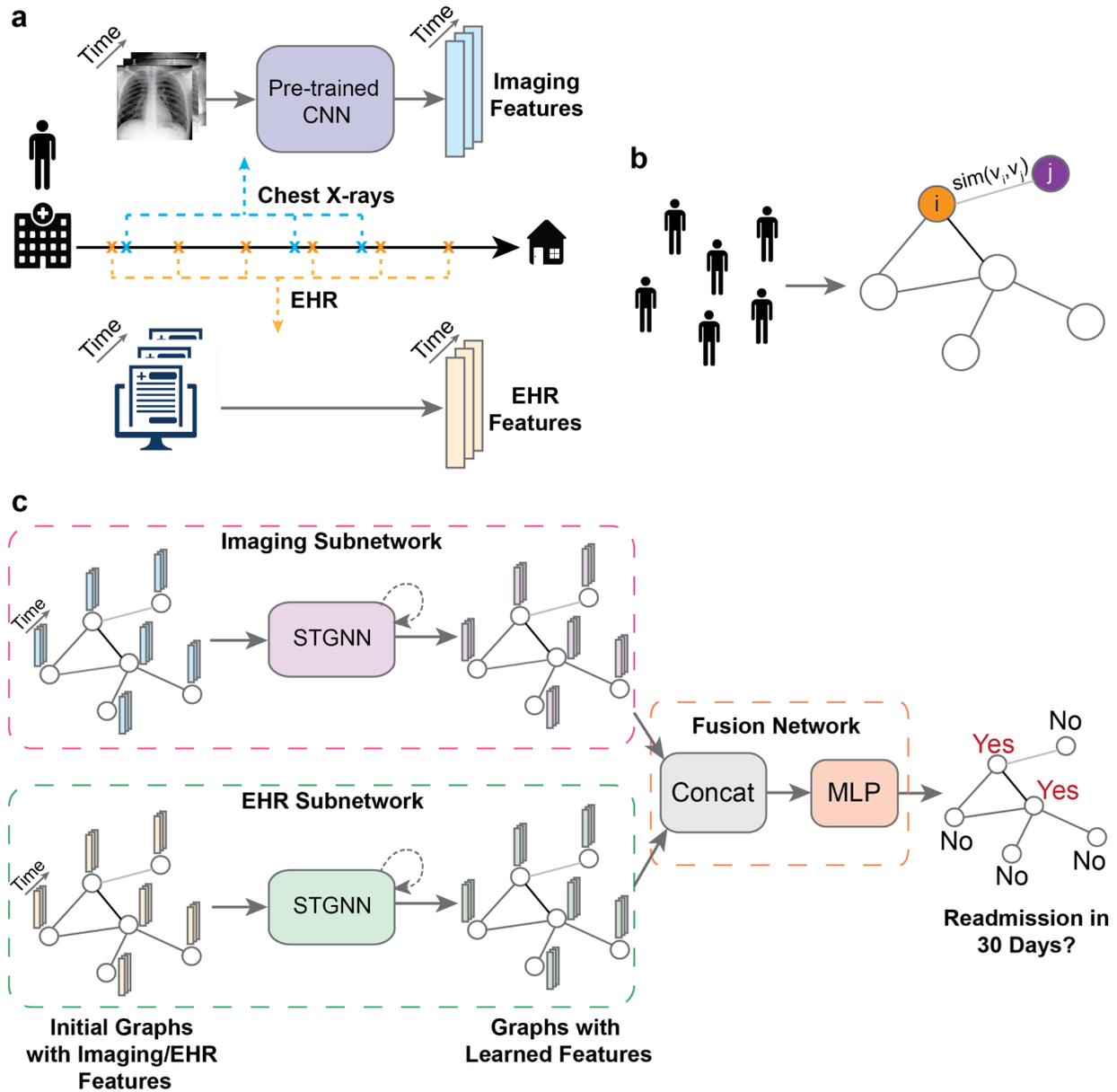

**Figure 1. Overview of our methods. (a)** We use patient longitudinal EHR and chest radiographs within hospitalization for 30-day all-cause hospital readmission prediction. **(b)** We represent hospital admissions as a graph, where each node corresponds to one admission, node features are longitudinal EHR or imaging features, and edges are represented by similarity between admissions measured by Euclidean distance between patient characteristics. **(c)** Architecture of our MM-STGNN model. *Abbreviations*: FC, fully connected; EHR, electronic health records; MLP, multilayer perceptron.

# Results

To leverage the topological relationship between hospital admissions and the temporal relationship in longitudinal hospital admission data, we frame the 30-day all-cause hospital readmission prediction problem as a node classification problem. Specifically, we represent hospital admissions as a graph, where each node in the graph corresponds to one hospital admission, the edges represent the similarity between the admissions defined by demographics or clinical variables, and the node features are longitudinal EHR or imaging (i.e., chest radiographs) features. To tackle the challenges regarding data heterogeneity and complexity of representing multimodal temporal data in a single space, we design a multimodal spatiotemporal graph neural network (MM-STGNN) to capture the topological structure between admissions and the temporal dependencies in longitudinal EHR and imaging data. Figure 1 illustrates our design. We demonstrate the effectiveness of our methods on two independent datasets: an internal dataset (referred to as "primary dataset" hereafter) as well as the public MIMIC-IV dataset[19,20]. Patient characteristics in both datasets are summarized in Table 1.

**Table 1. Patient characteristics in the primary dataset and MIMIC-IV dataset. Top three primary diagnoses are presented for the primary dataset, and top three comorbidities are presented for MIMIC-IV.**

| Characteristics | | Train | Validation | Test | Total |
|---|---|---|---|---|---|
| *Primary Dataset* | | | | | |
| **Age, years, mean (std)** | | 63.99 (19.01) | 64.01 (19.26) | 63.75 (19.45) | 63.94 (19.12) |
| **Gender, n (%)** | Female | 2911 (43.88%) | 321 (44.46%) | 795 (44.02%) | 4027 (43.95%) |
| | Male | 3708 (55.89%) | 399 (55.26%) | 1009 (55.87%) | 5116 (55.84%) |
| | Unknown | 15 (0.23%) | 2 (0.28%) | 2 (0.11%) | 19 (0.21%) |
| **Race, n (%)** | American Indian/Alaskan Native | 42 (0.63%) | 10 (1.39%) | 17 (0.94%) | 69 (0.75%) |
| | Asian | 103 (1.55%) | 15 (2.08%) | 35 (1.94%) | 153 (1.67%) |
| | Black | 254 (3.83%) | 24 (3.32%) | 61 (3.38%) | 339 (3.70%) |
| | Native Hawaiian/Pacific Islander | 8 (0.12%) | 1 (0.14%) | 7 (0.39%) | 16 (0.17%) |
| | White | 5491 (82.77%) | 611 (84.63%) | 1461 (80.90%) | 7563 (82.55%) |
| | Unknown | 736 (11.09%) | 61 (8.45%) | 225 (12.46%) | 1022 (11.15%) |
| **Ethnicity, n (%)** | Hispanic or Latino | 228 (3.44%) | 16 (2.22%) | 67 (3.71%) | 311 (3.39%) |
| | Not Hispanic or Latino | 5642 (85.05%) | 640 (88.64%) | 1533 (84.88%) | 7815 (85.30%) |
| | Unknown | 764 (11.52%) | 66 (9.14%) | 206 (11.41%) | 1036 (11.31%) |
| **Primary Diagnoses, n (%)** | Heart Disease | 3159 (43.78%) | 346 (43.69%) | 843 (43.21%) | 4348 (43.66%) |
| | Vascular Disease | 3026 (41.94%) | 368 (46.46%) | 825 (42.29%) | 4219 (42.37%) |
| | Metabolism Disease | 2076 (28.77%) | 219 (27.65%) | 548 (28.09%) | 2843 (28.55%) |
| **Length-of-stay, days, mean (std)** | | 11.15 (10.45) | 11.30 (11.70) | 11.11 (9.87) | 11.15 (10.45) |
| *MIMIC-IV Dataset* | | | | | |
| **Age, years, mean (std)** | | 66.29 (16.53) | 66.13 (16.91) | 65.83 (17.03) | 66.17 (16.69) |
| **Gender, n (%)** | Female | 3,367 (45.11%) | 865 (46.33%) | 1,070 (45.86%) | 5,302 (45.46%) |
| | Male | 4,097 (54.89%) | 1,002 (53.67%) | 1,263 (54.14%) | 6,362 (54.54%) |
| **Ethnicity, n (%)** | American Indian/Alaskan Native | 28 (0.38%) | 3 (0.16%) | 5 (0.21%) | 36 (0.31%) |
| | Asian | 231 (3.09%) | 66 (3.54%) | 85 (3.64%) | 382 (3.28%) |
| | Black/African Black | 875 (11.72%) | 203 (10.87%) | 259 (11.10%) | 1,337 (11.46%) |
| | Hispanic/Latino | 305 (4.09%) | 70 (3.75%) | 109 (4.67%) | 484 (4.15%) |
| | White | 5,113 (68.50%) | 1,306 (69.95%) | 1,592 (68.24%) | 8,011 (68.68%) |
| | Other | 354 (4.74%) | 84 (4.50%) | 96 (4.11%) | 534 (4.58%) |
| | Unknown | 558 (7.48%) | 135 (7.23%) | 187 (8.02%) | 880 (7.54%) |
| **Comorbidity, n (%)** | Diseases of the Circulatory System | 1,238 (13.35%) | 301 (12.95%) | 356 (12.13%) | 1,895 (13.04%) |
| | Diseases of the Immune Mechanism | 475 (5.12%) | 112 (4.82%) | 165 (5.62%) | 752 (5.17%) |
| | Endocrine, Nutritional and Metabolic Diseases | 342 (3.69%) | 81 (3.48%) | 117 (3.99%) | 540 (3.72%) |
| **Length-of-stay, days, mean (std)** | | 14.98 (18.47) | 14.75 (12.52) | 15.69 (17.07) | 15.09 (17.37) |

**Prediction of 30-day all-cause hospital readmission**

To investigate the effectiveness of our proposed MM-STGNN, we compare our models to the following baselines: (a) XGBoost[21], a state-of-the-art gradient-boosting algorithm, (b) GraphSAGE[22], for comparison with non-temporal GNN, and (c) Long Short-Term Memory network[23] (LSTM), for comparison with traditional temporal model. For non-temporal models such as XGBoost and GraphSAGE, imaging or EHR features in the last time step are used (see Methods "node features" section).

Table 2 (third and fourth columns) shows the performance of MM-STGNN and the baselines, where procedures performed during the hospital stay (i.e., CPT codes) are used to construct the admission graph for the primary dataset, and patient demographics (age, gender, and ethnicity) are used to construct the admission graph for MIMIC-IV. MM-STGNN achieves 0.788 and 0.791 AUROC on the primary and MIMIC-IV datasets, respectively, significantly outperforming the baselines (Delong's $p < 0.05$).

Figure 2a (first and second columns) shows the ROC curves of MM-STGNN on the primary and MIMIC-IV datasets. At a 80% sensitivity, MM-STGNN achieves a specificity of 60%, which translates clinically to missing 20% readmissions with 40% predicted readmissions being false positives. On MIMIC-IV, MM-STGNN achieves a specificity of 54% at a 80% sensitivity; this translates to missing 20% readmissions with 46% predicted readmissions being false positives. Figure 2b (first and second columns) shows the precision-recall (PR) curves of MM-STGNN. MM-STGNN has an average precision of 64% on both datasets.

**Table 2. 30-day all-cause readmission prediction results for baselines and MM-STGNN on the primary and MIMIC-IV datasets, as well as a subset in the primary test set where LACE+ scores are available (n = 255)**. Confidence intervals (CI) are calculated using the DeLong method[24]. Best results without statistically significant differences are highlighted in bold.

| Model | Modality | Primary Dataset AUROC [95% CI] | MIMIC-IV AUROC [95% CI] | Primary Dataset LACE+ Subset AUROC [95% CI] |
|---|---|---|---|---|
| LACE+ | EHR | - | - | 0.614 [0.542-0.686] |
| XGBoost | Imaging | 0.663 [0.636-0.691] | 0.689 [0.664-0.714] | 0.607 [0.529-0.685] |
| GNN | Imaging | 0.697 [0.671-0.723] | 0.631 [0.605-0.656] | 0.617 [0.540-0.694] |
| LSTM | Imaging | 0.689 [0.663-0.715] | 0.709 [0.682-0.736] | 0.661 [0.587-0.735] |
| XGBoost | EHR | 0.738 [0.714-0.763] | 0.689 [0.664-0.714] | **0.700 [0.630-0.770]** |
| GNN | EHR | 0.723 [0.697-0.749] | 0.720 [0.694-0.745] | 0.700 [0.627-0.773] |
| LSTM | EHR | 0.761 [0.738-0.785] | 0.730 [0.705-0.755] | **0.785 [0.726-0.844]** |
| MM-STGNN | EHR+Imaging | **0.788 [0.765-0.811]** | **0.791 [0.766-0.816]** | **0.772 [0.709-0.834]** |

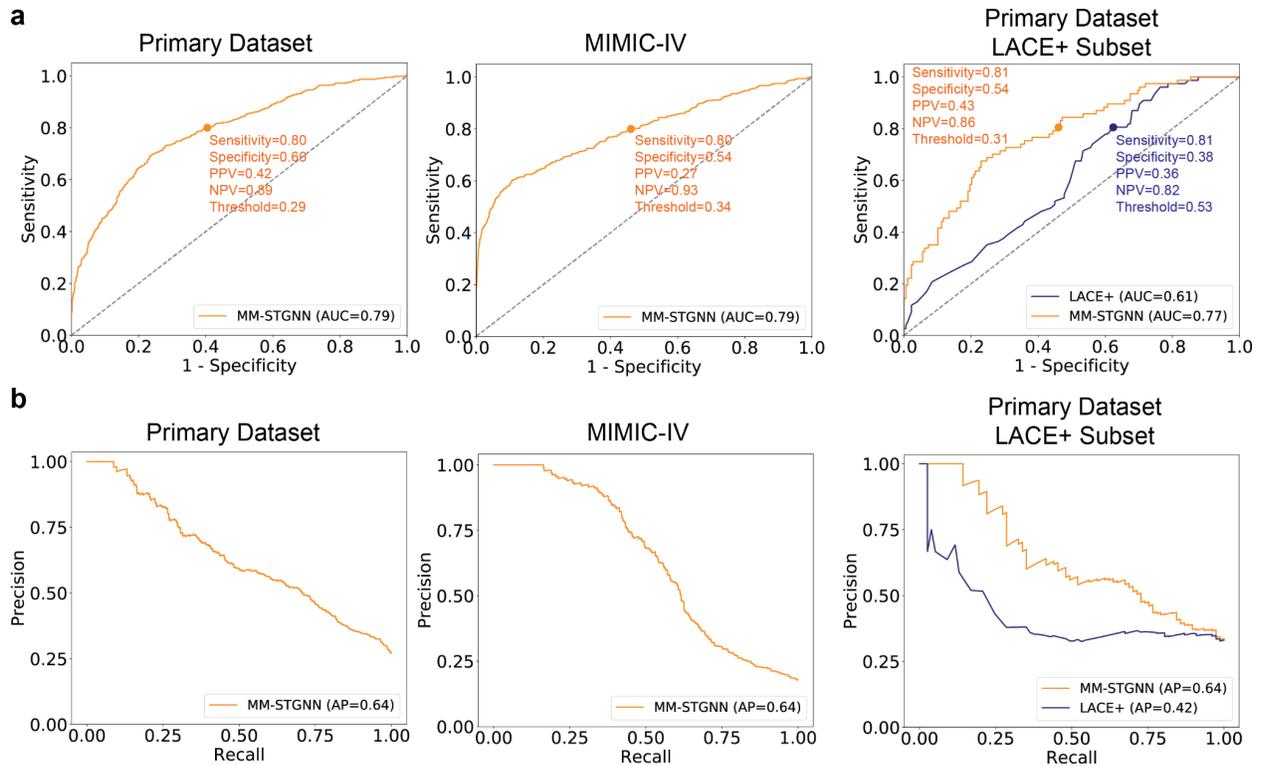

**Figure 2. (a) Receiver operating characteristics curves of MM-STGNN.** Dots represent the operating points at 80% sensitivity. Texts show the sensitivity, specificity, positive predictive value (PPV), negative predictive value (NPV), and the probability threshold for the respective model. **(b) Precision-recall curves of MM-STGNN.**

**Comparison to LACE+**

To compare our models to the clinically used LACE+ scores[3], we evaluate our model performance on a subset of admissions in the primary test set where LACE+ scores are available (n = 255). As LACE+ scores range from 0 to 90 with higher scores indicating higher risks of readmission, we treat LACE+ scores as probabilities of readmission and compare them to our model predictions. As shown in Table 2 (last column), MM-STGNN significantly outperforms LACE+ score, providing 15.8 points improvement in AUROC. Figure 2a (last column) shows the ROC curves of LACE+ and MM-STGNN on the primary test subset where LACE+ scores are available. At a sensitivity of 81%, LACE+ has a specificity of 38%, which translates to missing 19% readmissions with 62% predicted readmissions being false positives. In contrast, MM-STGNN has a specificity of 54% at 81% sensitivity, a 16 points increase in specificity compared to LACE+. This suggests that at the same sensitivity, MM-STGNN results in fewer false positives than LACE+.

Figure 2b (last column) shows the PR curves of LACE+ and MM-STGNN on the primary test subset where LACE+ scores are available. MM-STGNN achieves an average precision of 0.64, a 22 points improvement over LACE+. These results demonstrate that our MM-STGNN is better at predicting 30-day all-cause readmission while reducing false positives than the clinically used LACE+ score.

**Prediction of all-cause hospital readmission on disease-specific subpopulations**

To examine the model performance for disease-specific subpopulations, we utilize the primary diagnoses processed from clinical notes to group admissions in the primary dataset into subsets (see Methods "primary dataset" section), where admissions in each subset are associated with the same primary diagnosis. Note that these subsets are not mutually exclusive, i.e., one admission may be associated with more than one primary diagnosis. Table 3 shows the model performance evaluated on the three largest disease-specific subsets in the primary test set. MM-STGNN outperforms the baselines on patients with heart diseases and vascular diseases.

For the MIMIC-IV dataset where clinical notes are not available, we use ICD-10 codes to group admissions into disease-specific subsets. Supplementary Table 1 shows the model prediction performance on the three largest subsets in MIMIC-IV (i.e., diseases of circulatory system, diseases of immune mechanism, and endocrine, nutritional and metabolic diseases). We observe that MM-STGNN outperforms the baselines on patients with circulatory diseases and endocrine, nutritional and metabolic diseases.

**Table 3.** 30-day all-cause readmission prediction results for baselines, STGNN, and MM-STGNN on disease-specific subsets in the primary test set. Confidence intervals (CI) are calculated using the DeLong method[24]. Best results without statistically significant differences are highlighted in bold.

| Model | Modality | Heart Disease (n = 843) AUROC [95% CI] | Vascular Disease (n = 825) AUROC [95% CI] | Metabolism Disease (n = 548) AUROC [95% CI] |
|---|---|---|---|---|
| XGBoost | Imaging | 0.687 [0.646-0.727] | 0.664 [0.622-0.706] | 0.663 [0.614-0.712] |
| GNN | Imaging | 0.724 [0.687-0.762] | 0.700 [0.662-0.739] | 0.682 [0.635-0.728] |
| LSTM | Imaging | 0.706 [0.668-0.744] | 0.690 [0.650-0.729] | 0.655 [0.607-0.703] |
| XGBoost | EHR | 0.748 [0.711-0.785] | 0.721 [0.683-0.759] | **0.721 [0.676-0.766]** |
| GNN | EHR | 0.736 [0.697-0.775] | 0.731 [0.693-0.770] | 0.679 [0.631-0.727] |
| LSTM | EHR | 0.763 [0.728-0.797] | **0.759 [0.724-0.795]** | **0.724 [0.679-0.769]** |
| MM-STGNN | EHR+Imaging | **0.800 [0.765-0.834]** | 0.786 [0.751-0.821] | 0.747 [0.703-0.792] |

**Interpretation of node features and neighbors**

To understand the importance of each node's imaging and EHR features and its neighboring nodes, we perform interpretability analysis by extending GNNExplainer[25] to our MM-STGNN fusion framework (see Methods). Briefly, given a node-of-interest, GNNExplainer identifies a subgraph structure among the node's k-hop neighbors and a subset of its node features that are crucial to the GNN's prediction of the node[25].

Figure 3a shows the interpretation of a node (node ID 1611) that is correctly predicted as positive by both LACE+ and MM-STGNN, where the decision threshold is determined by the Youden J's statistic[26,27]. The top left panel shows the center node's primary diagnoses, demographics, length-of-stay, and LACE+ score; the top right panel shows the importance of the center node's 1-hop neighbors, where the importance scores are indicated on the edges and the most important neighboring node's primary diagnoses are listed (i.e., node ID 1386); the bottom panel shows the feature importance of the 100 most important features, where the imaging feature importance is averaged. As shown in Figure 3a, while primary diagnoses were not explicitly used to train MM-STGNN, the most important node features reflect the patient's primary diagnoses (bottom panel). For example, one of the patient's primary diagnoses is "viral infectious disease", which is reflected in the selected node features "sodium, U" (laboratory results). Additionally, the most important neighbor's primary diagnoses are the same as the center node (top right panel), which likely makes it the most crucial neighbor for the center node's prediction.

Figure 3b shows the interpretation of a node (node ID 3370) that is a false positive of LACE+ but true negative of MM-STGNN. Similar to Figure 3a, the most important node features reflect the primary diagnoses of the patient (bottom panel of Figure 3b); the most important neighbor node's label and two primary diagnoses are the same as the center node (top right panel). In contrast, the patient has a LACE+ score of 72, which incorrectly predicts the center node as positive (i.e., readmitted within 30 days).

In Figure 3c, we show the explanation of a node (node ID 754) that is a false positive of MM-STGNN. Given the patient's current status of long hospital stay, primary brain disease, history of mental health disorder, hyperintensive disease and anticoagulation medication intake, the model identifies a cancer patient with cardiovascular disease as the most important neighbor and predicts positive for 30-day readmission following the same trend of the neighbor. This incorrect prediction could also be influenced by the rarity of mental health patients within the primary cohort.

Figure 3d illustrates an example of a node (node ID 2525) that is a false negative of MM-STGNN where the patient is relatively young (41 years old) and healthy (no prior chronic disease history), and admitted for bacterial pneumonia. The model identifies normal laboratory test results (i.e., sodium, creatinine, troponine, anion gap) as top features, which could be the reason why the model finally results in a false negative prediction for this node.

## a. Explaination of node 1611

**Primary Diagnoses:**
- Vascular disease
- Heart disease
- Viral infectious disease

**Demographics:**
- Age: 39
- Gender: Female
- Ethnicity: Not hispanic or latino
- Race: Black

**Length-of-Stay:** 34 days

**LACE+ Score:** 72

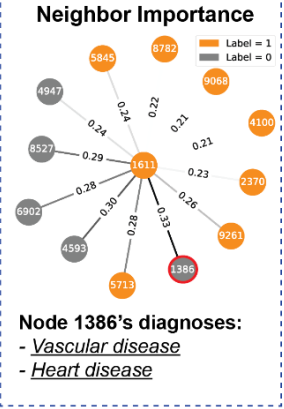

**Node 1386's diagnoses:**
- *Vascular disease*
- *Heart disease*

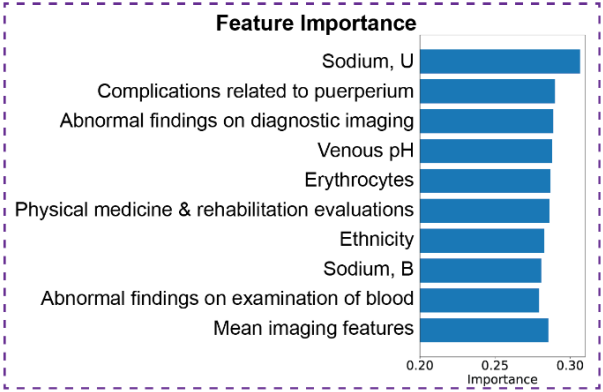

## b. Explaination of node 3370

**Primary Diagnoses:**
- Vascular disease
- Heart disease
- Disease of metabolism
- Blood coagulation disease

**Demographics:**
- Age: 54
- Gender: Female
- Ethnicity: Not hispanic or latino
- Race: White

**Length-of-Stay:** 6 days

**LACE+ Score:** 72

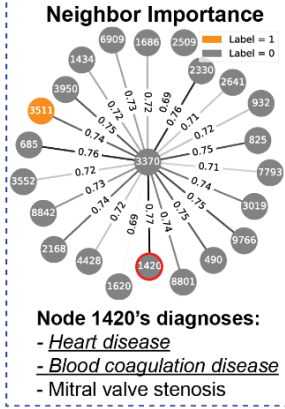

**Node 1420's diagnoses:**
- *Heart disease*
- *Blood coagulation disease*
- Mitral valve stenosis

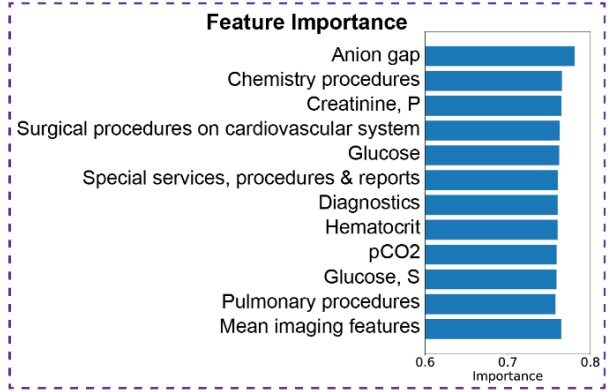

## c. Explaination of node 754

**Primary Diagnosis:**
- Brain disease

**Demographics:**
- Age: 70
- Gender: Male
- Ethnicity: Hispanic or latino
- Race: Unknown

**Length-of-Stay:** 19 days

**LACE+ Score:** Not available

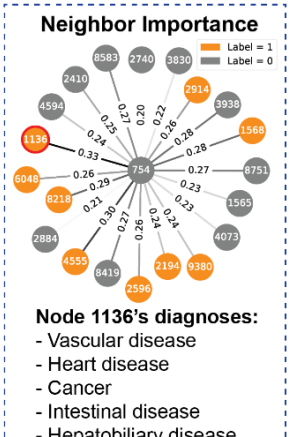

**Node 1136's diagnoses:**
- Vascular disease
- Heart disease
- Cancer
- Intestinal disease
- Hepatobiliary disease

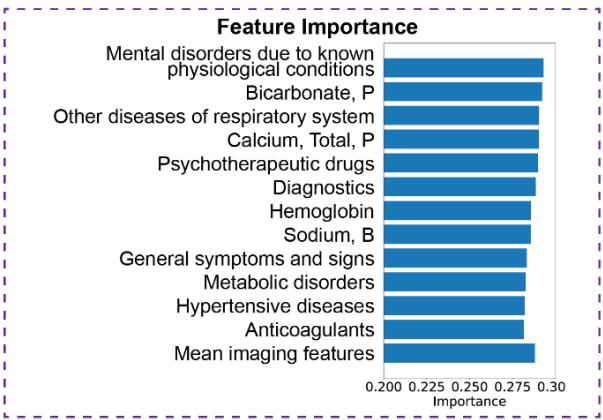

## d. Explaination of node 2525

**Primary Diagnosis:**
- Bacterial pneumonia

**Demographics:**
- Age: 41
- Gender: Female
- Ethnicity: Not hispanic or latino
- Race: Black

**Length-of-Stay:** 2 days

**LACE+ Score:** Not available

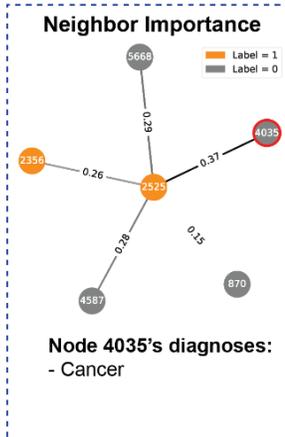

**Node 4035's diagnoses:**
- Cancer

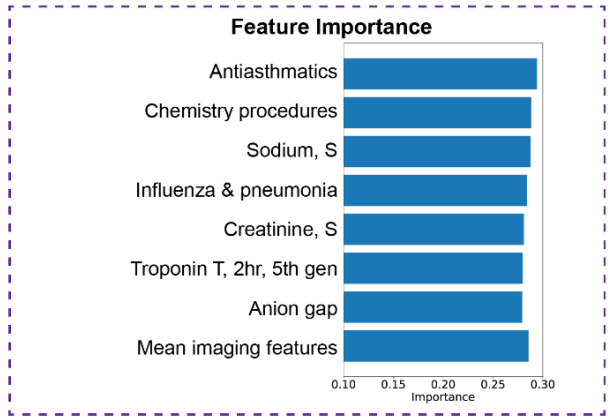

**Figure 3. Example interpretability of (a) a node that is true positives of both LACE+ and MM-STGNN, (b) a node that is a false positive of LACE+ but true negative of MM-STGNN, (c) a node that is a false positive of MM-STGNN, and (d) a node that is a false negative of MM-STGNN.** In each subfigure, the top left panel shows the center node's primary diagnoses, demographics, length-of-stay, and LACE+ score; the top right panel shows the importance of the center node's 1-hop neighbors, where the importance scores are indicated on the edges and the most important neighboring node's primary diagnoses are listed; the bottom panel shows the feature importance of the 100 most important EHR and imaging features, with the imaging feature importance averaged (90 imaging features in 3a, 89 imaging features in 3b, 88 imaging features in 3c, and 93 imaging features in 3d) and shown in the last row. Orange nodes have positive labels (i.e., readmitted within 30 days), while gray nodes have negative labels. Numbers on the nodes indicate node IDs.

**Effects of node feature modalities**

To examine the individual contributions of imaging and EHR node features, we perform an ablation study by training imaging-based and EHR-based STGNN subnetworks in MM-STGNN independently for 30-day all-cause readmission prediction on both datasets. As shown in Supplementary Table 2, we observe that the performance gain of MM-STGNN mainly comes from EHR node features; this could be due to the fact that the extracted chest radiograph features may not be as informative as EHR features for predicting all-cause readmission on a generic population. We also evaluated the trained models on disease-specific subgroups in the primary dataset (i.e., heart disease, vascular disease, and metabolic disease), and we find that performance gain of chest radiographs is limited for all the major disease subgroups (Supplementary Table 3).

**Effects of temporarity**

To examine the effects of temporality, we conduct an ablation study by training a multimodal GNN (MM-GNN) using the last available imaging and EHR features. As shown in Supplementary Table 4, MM-STGNN with temporal data achieves significantly better performance than MM-GNN without temporal data on the primary dataset and MIMIC-IV (Delong's $p < 0.001$). This indicates the importance of leveraging in-patient temporal data for more accurate predictions.

**Effects of EHR edge features on model performance**

To examine the effects of different EHR sources used in graph construction (see Methods), we compare the performance of MM-STGNN using different EHR sources in the edges (e.g., procedure, diagnosis, labs). In Supplementary Table 5, we observe that different EHR sources result in similar model performance, suggesting that they are highly correlated and often represent similar patient status.

## Discussion

In this study, we propose a graph-based modeling approach for 30-day all-cause hospital readmission. Our approach captures the topological relationships between hospital admissions and the temporal relationships in patient longitudinal EHR and imaging data. By combining both EHR and imaging data, our MM-STGNN provides an AUROC of 0.79 on both primary and MIMIC-IV datasets. Importantly, MM-STGNN significantly outperforms the current clinical reference standard (LACE+) on 30-day all-cause readmission prediction on the primary dataset. Additionally, we show that our MM-STGNN outperforms the baselines in patients with heart diseases and vascular diseases. Lastly, our qualitative model interpretability analysis suggests that a node's most important features reflect the patient's primary diagnoses.

Our methods have several important advantages. First, we show substantially improved 30-day all-cause prediction performance, achieving AUROC values of 0.79 on two datasets from independent medical centers. Prior studies on all-cause hospital readmission prediction report a wide range of model performance, with AUROC values varying from 0.64 to 0.90[7–10,13,15,16,18]. However, only one study has validated their methods on two independent datasets and achieved 0.75–0.76 AUROCs[13]. Furthermore, our MM-STGNN significantly outperforms the clinically used LACE+ scores, providing 15.8 points improvement in AUROC (Table 2). It is important to note that LACE+ score utilizes the patient's medical histories[3] (e.g., visits to the emergency department in the past 6 months). In contrast, our methods only use the patient's data during the current hospital stay and do not consider previous emergency department visits or other hospital stays. Therefore, our methods can be used for patients without well-documented medical histories and could avoid the potential bias of prior admissions, which provides more flexibility than LACE+ scores.

Second, our models are also better at predicting 30-day hospital readmission on disease-specific subpopulations. Notably, MM-STGNN achieves an AUROC of 0.80 on patients with heart diseases in the primary dataset. In contrast, previous studies on patients with congestive heart failure report AUROCs of 0.64[7,11] and 0.77[17]. The ability to accurately predict 30-day readmission on patients with certain conditions would be valuable for identifying targeted interventions to prevent unnecessary readmissions.

Third, our MM-STGNN architecture is agnostic to node feature modalities. While we demonstrate our methods on EHR and imaging data, additional modalities—such as electrocardiograms, vital signs, and clinical notes—can be easily included in MM-STGNN. In contrast, clinically-used risk scores, such as LACE+ scores, are based on a small, fixed number of clinical variables[3].

Fourth, our model interpretability method provides intuitive explanations of the model predictions. By visualizing the most crucial features and similar patients used by the model, physicians may better understand the intuition of features used by the model as they make decisions regarding disposition.

Finally, our study has important clinical value. Our model could be used to more accurately aid in decisions regarding discharge. More specifically, the model could help identify at-risk patients for readmission, thereby allowing healthcare facilities to triage resources accordingly, including follow-up phone calls to patients, social work follow-up, and follow-up visits with providers. Finally, decisions regarding discharge disposition location (i.e., home versus skilled facility care) could also be tailored using information provided by our model.

There are several limitations in this study, such as those associated with a retrospective design. First, we treat one chest radiograph as one time step, which ignores the fact that the chest radiograph studies were performed on irregular intervals (e.g., first day and third day of admission). This may explain why the performance gain from chest radiographs is limited (Supplementary Table 2 and Supplementary Table 3). In the future, an improved approach that takes into account the irregular intervals of chest radiographs is needed to more accurately capture the temporal dependencies. Second, due to the lack of clinical notes in MIMIC-IV, we use ICD-10 codes instead of clinical notes to identify disease-specific subpopulations, and thus a direct comparison between the primary dataset and MIMIC-IV for model performance on disease-specific subsets is limited. Third, MIMIC-IV dataset does not have the same set of clinical variables as the primary dataset, particularly for procedure billing codes. Thus, we did not directly evaluate MM-STGNN trained on the primary dataset on the MIMIC-IV data. Future work that projects heterogeneous EHR data to the same embedding space is needed to allow direct evaluation of trained models on independent validation datasets. Lastly, while we consider patients who died within hospitalization as readmitted within 30 days, future studies on multiclass classification of 30-day hospital readmission versus mortality are needed to distinguish these classes.

In conclusion, we propose a multimodal spatiotemporal graph neural network for 30-day all-cause hospital readmission. Our approaches outperform the baselines and the traditional, clinically used LACE+ scores on general patient populations, provide superior performance than the baselines on disease-specific subpopulations, and offer intuitive explanation of the model predictions.

# Methods

**Datasets**

*Primary dataset*

With the approval of the local Institutional Review Board (IRB), we collected an in-house de-identified dataset of 44,936 hospital admissions from 35,260 unique patients between January 1, 2019 and December 31, 2019. We retrieved comprehensive EHR and chest radiographs acquired during these hospital stays. Data inclusion criteria are as follows: (a) hospital stay $\geq$ 48 hours, (b) at least 2 chest radiographs acquired within the hospital stay, (c) admission types that are not ambulatory observation, EU observation, or direct observation, (d) discharge locations that are not acute hospital, healthcare facilities, or against advice. Data exclusion criteria are (a) chest radiographs studies with only CLAHE images and (b) chest radiographs without anterior-posterior or posterior-anterior chest views. Following the inclusion and exclusion criteria, a total of 9,958 hospitalizations with 44,084 chest radiographs from 9,162 unique patients are included in the analysis. Given continuous records of hospital admission and discharge, we are able to identify 2,675 instances of hospital admissions where the patients were readmitted within 30 days (inclusive). Patients who died during hospitalization or within 30 days after discharge are considered as positive for readmission. Supplementary Figure 1a illustrates the detailed inclusion and exclusion criteria and Table 1 shows the patient characteristics. Using standard rule-based natural language processing techniques, we parse free-text discharge notes and extract the primary diagnoses or reasons for admission. We then map these texts to Human Disease Ontology[28] and identify 189 unique diagnosis classes in our cohort by mapping leaf nodes of the ontology to mid-level nodes. These identified primary diagnoses are used to group patients into disease-specific subsets for model evaluation (see Results). Supplementary Table 6 lists the 50 most frequent primary diagnoses in our cohort.

We collected LACE+ scores[3] (length of stay, acuity, comorbidities, emergency room visits index) from a subset of internal patients (n = 1307) where the LACE+ scores were calculated as a standard procedure in the clinical setting. LACE+ is used to predict the risk of post-discharge death or urgent readmission on the basis of administrative data. The index is calculated using a pre-selected list of covariates (e.g. age, gender, acuity of admission) and the model was originally developed based on administrative data in Ontario, Canada[3].

*External dataset*

We use MIMIC-IV v1.0 "hosp" module[19,20] as the external dataset, where patient demographics are obtained from MIMIC-IV v1.0 "core" module and chest radiographs are obtained from MIMIC-CXR-JPG v2.0.0[20,29,30]. Data inclusion and exclusion criteria are similar to the internal cohort. Following the inclusion and exclusion criteria, a total of 14,532 hospital admissions with 87,472 chest radiographs from 11,664 unique patients are left, where patients from 2,552 admissions were readmitted within 30 days of discharge (inclusive). Patients who died during hospitalization are considered as positive for readmission. Supplementary Figure 1b shows the detailed inclusion and exclusion criteria and Table 1 shows patient characteristics for MIMIC-IV.

**Admission graph structure**

We hypothesize that hospital admissions with similar patient demographics (e.g., age and gender) or clinical profile (e.g., procedures performed, comorbidities, and lab results) would have similar post-discharge outcomes, and thus should have stronger connections. Therefore, we compute the edge weight $W_{ij}$ between admission node $i$ and $j$ as: $W_{ij} = exp(-\frac{dist(v_i, v_j)^2}{\sigma^2})$ $if$ $W_{ij} \geq threshold$, $else$ $W_{ij} = 0$. Here, $v_i$ and $v_j$ are EHR features for node $i$ and node $j$, respectively, $dist(v_i, v_j)$ is the Euclidean distance between $v_i$ and $v_j$, $\sigma$ is the standard deviation of the distances, and $W_{ij} \in [0, 1]$. To introduce graph sparsity, the threshold is determined by keeping the top κ% edges in the graph, where κ is a hyperparameter. Note that the Gaussian kernel results in a larger $W_{ij}$ for a smaller $dist(v_i, v_j)$ (i.e., similar nodes have larger edge weights), and vice versa, which is a commonly used approach when the edge weights are not naturally defined[31]. Figure 1b illustrates the admission graph construction.

**Node features**

*EHR node features for primary dataset*

For the primary dataset, EHR data collected for patients include (a) demographics (i.e., age, gender, race, and ethnicity), (b) procedures recorded as CPT codes, (c) comorbidities recorded as ICD-10 codes, (d) laboratory test results, and (e) medications administered during hospital stays from billing records. First, following a prior study[32], we utilize hierarchical structure of CPT and ICD-10 codes to map individual codes to their respective subgroups, resulting in 72 CPT and 243 ICD-10 subgroups. Second, clinical experts select 51 most important laboratory tests. We one-hot encode the laboratory test values as "normal" or "abnormal" based on known ranges of normal results. Third, medications are mapped to their RxNorm therapeutic classes, resulting in 45 unique medication classes. Lastly, to leverage temporal relationship in EHR data, we treat one day within hospitalization as one time step, where the EHR feature vector at each time step is the concatenation of (a) counts of CPT/ICD-10 subgroups as well as medications within that day, (b) one-hot encoded lab results within that day, and (c) patient demographics. Because the hospital admissions have different lengths-of-stay, we only use the last $T_{EHR}$-day EHR features and pad admissions shorter than $T_{EHR}$ days with the last available EHR features. In our experiments, we treat $T_{EHR}$ as a hyperparameter. For each categorical EHR feature (e.g., gender and lab results), we use trainable embeddings with dimension $D_{cat}$ to map to continuous values, where $D_{cat}$ is a hyperparameter.

*EHR node features for external dataset*

For the MIMIC-IV dataset, we select the same set of EHR data as the primary dataset whenever possible. EHR data include (a) demographics (age, gender, and ethnicity), (b) comorbidities recorded as ICD-10 codes, (c) laboratory test results, and (d) medications administered during hospitalizations. Procedural CPT codes are not included because there are only 179 records in our cohort. We follow the same EHR preprocessing steps detailed above, resulting in 91 unique ICD-10 subgroups, 36 unique laboratory tests, and 41 unique RxNorm therapeutic classes for medications. Similar to the primary dataset, we map categorical EHR features to continuous values using trainable embeddings with dimension $D_{cat}$.

*Imaging node features*

For both primary and MIMIC-IV datasets, imaging features are extracted from a DenseNet121[33], pretrained using MoCo[34] self-supervised pretraining protocol on CheXpert dataset[35,36]. We choose this pretrained model because the representations learned in MoCo self-supervised pretraining has been shown to generalize well to a variety of downstream tasks[34]. To leverage the temporal relationship in chest radiographs, we treat each chest radiograph as one time step. To handle variable number of chest radiographs across admissions, we only use the last $T_{CXR}$ imaging features and pad the short sequences with the last available imaging features. Similar to EHR features, we treat $T_{CXR}$ as a hyperparameter in our experiments.

**Features for edges**

To capture the topological relationship between hospital admissions, we use the EHR features to construct distinct admission graphs (Section "Admission graph structure"). We examine the effect of different EHR sources (i.e., demographics, CPT, ICD-10, lab results, and medications) on the model performance.

**Spatiotemporal graph neural network**

To model the spatiotemporal dependencies in patient longitudinal imaging or EHR data, we propose a spatiotemporal graph neural network (STGNN) by replacing the matrix multiplications in Gated Recurrent Units[37] (GRU) with GraphSAGE[22] layers. Specifically, let $X^{(t)} \in R^{N \times D}$ be the EHR or imaging node features at time step $t$ with $N$ nodes and $D$ feature dimension, the outputs of a STGNN cell at time step $t$ are computed as follows:

$$r^{(t)} = \sigma(\Theta_{r \star G}[X^{(t)}, H^{(t-1)}] + b_r), \quad u^{(t)} = \sigma(\Theta_{u \star G}[X^{(t)}, H^{(t-1)}] + b_u)$$

$$C^{(t)} = tanh(\Theta_{C \star G}[X^{(t)}, (r^{(t)} \odot H^{(t-1)})] + b_C), \quad H^{(t)} = u^{(t)} \odot H^{(t-1)} + (1 - u^{(t)}) \odot C^{(t)}$$

Here, $H^{(t)}$ denotes the output of STGNN cell at time step $t$, $\sigma$ denotes Sigmoid function, $\odot$ represents the Hadamard product, $r^{(t)}$, $u^{(t)}$, $C^{(t)}$ denote reset gate, update gate and candidate at time step $t$, respectively, $\star G$ denotes the GraphSAGE operation, $\Theta_r$, $b_r$, $\Theta_u$, $b_u$, $\Theta_C$ and $b_C$ are the weights and biases for the corresponding GraphSAGE layers. Note that while we choose GraphSAGE due to its inductive bias and GRU due to its improved long-term memory, our method works for any types of GNN and recurrent neural network layers.

**Multimodal fusion**

To leverage multiple data modalities, we propose a multimodal fusion framework that consists of (a) two separate STGNN subnetworks for each modality that map imaging and EHR features to the same hidden dimension $D_{hidden}$ and (b) a fusion network (i.e., multi-layer perceptron) that takes as inputs the concatenated representations from the imaging and EHR subnetworks and produces a prediction for each node in the admission graph. Note that the graphs in the subnetworks have different node features (i.e., imaging or EHR) but the same edges.

**Model interpretability**

To understand the importance of each node's imaging/EHR features and neighboring nodes, we perform interpretability analysis by extending GNNExplainer[25] to our multimodal fusion framework. To explain the importance of imaging and EHR node features for MM-STGNN, we use two separate node feature masks for each subnetwork in MM-STGNN; to explain the importance of a node's neighbors, we use only one edge mask because the graph edges are the same in both subnetworks in our study. For details about GNNExplainer, we refer the readers to the original paper[25]. This interpretability analysis allows us to examine the important imaging and EHR node features, as well as the crucial neighboring nodes for MM-STGNN's prediction of a node-of-interest.

**Model training procedure**

Training for all models was accomplished using the Adam optimizer[38] and cosine annealing learning rate scheduler[39] (without warm start) in PyTorch on a single NVIDIA RTX GPU. We performed the following hyperparameter search on the validation set: (a) initial learning rate within range [1e-5, 1e-2]; (b) the number of STGNN layers within range {1, 2} and hidden units $D_{hidden}$ within range {64, 128, 256}; (c) dropout probability in the last fully connected layer within range [0, 0.5]; (d) graph sparsity hyperparameter κ within range [1e-5, 10]; (e) EHR sequence length $T_{EHR}$ and imaging sequence length $T_{CXR}$ within range [3, 15]; (f) categorical feature embedding dimension $D_{cat}$ within range {1,2,3}; (g) fusion network multi-layer perceptron hidden units within range {128, 256}. The final hyperparameters were selected based on the highest AUROC on the validation set. In all experiments, the model was trained for 100 epochs and was stopped early when the validation loss did not decrease for 10 consecutive epochs.

**Model evaluation strategy and baselines**

Datasets are randomly split based on unique patients with a ratio of 80%/20%, where 20% of the patients are reserved as a held-out test set, and the remaining 80% are further split into train and validation sets for model training and hyperparameter tuning, respectively. The train/validation set ratio is approximately 72%/8% for the primary dataset and 65%/16% for MIMIC-IV. Baseline models include: (a) XGBoost[21], (b) GraphSAGE[22], and (c) Long Short-Term Memory network[23] (LSTM). Moreover, we compare our model performance to the test subset in the primary dataset where LACE+ scores are available.

We use AUROC as the primary metric and perform DeLong non-parametric tests[24] to evaluate the equivalence of AUROC values. A statistical significance (α) threshold of 0.05 is used for all reported tests.

## Data availability

The primary dataset used in this study is not publicly available because of Protected Health Information restrictions. The external MIMIC-IV dataset is publicly available at https://physionet.org/content/mimiciv/1.0/. The pretrained DenseNet model used for extracting imaging features is publicly available at https://github.com/stanfordmlgroup/MoCo-CXR. The code used in this study for the MIMIC-IV dataset will be made publicly available at https://github.com/tsy935/readmit-stgnn.

# Declaration of interests

The authors report no conflict of interests.

## Author contributions

## Acknowledgements

The work is partially supported by Mayo Clinic CPC 2022 Clinical Practice Innovation Program (Company/PAU/Activity: 301/44708/22INAREA).

# References

1. McIlvennan, C. K., Eapen, Z. J. & Allen, L. A. Hospital readmissions reduction program. *Circulation* **131**, 1796–1803 (2015).

2. America's Health Rankings analysis of U.S. HHS, Centers for Medicare & Medicaid Services, Office of Minority Health, Mapping Medicare Disparities Tool, United Health Foundation, AmericasHealthRankings.org. Hospital Readmissions - Ages 65-74. https://www.americashealthrankings.org/explore/senior/measure/hospital_readmissions_sr/state/U.S.

3. van Walraven, C., Wong, J. & Forster, A. J. LACE+ index: extension of a validated index to predict early death or urgent readmission after hospital discharge using administrative data. *Open Med.* **6**, e80–90 (2012).

4. Donzé, J., Aujesky, D., Williams, D. & Schnipper, J. L. Potentially avoidable 30-day hospital readmissions in medical patients: derivation and validation of a prediction model. *JAMA Intern. Med.* **173**, 632–638 (2013).

5. Donzé, J. D. *et al.* International Validity of the HOSPITAL Score to Predict 30-Day Potentially Avoidable Hospital Readmissions. *JAMA Intern. Med.* **176**, 496–502 (2016).

6. Morgan, D. J. *et al.* Assessment of Machine Learning vs Standard Prediction Rules for Predicting Hospital Readmissions. *JAMA Netw Open* **2**, e190348 (2019).

7. Allam, A., Nagy, M., Thoma, G. & Krauthammer, M. Neural networks versus Logistic regression for 30 days all-cause readmission prediction. *Sci. Rep.* **9**, 9277 (2019).

8. Huang, K., Altosaar, J. & Ranganath, R. ClinicalBERT: Modeling Clinical Notes and Predicting Hospital Readmission. *arXiv [cs.CL]* (2019).

9. Jiang, S., Chin, K.-S., Qu, G. & Tsui, K. L. An integrated machine learning framework for hospital readmission prediction. *Knowledge-Based Systems* **146**, 73–90 (2018).

10. Li, Q., Yao, X. & Échevin, D. How Good Is Machine Learning in Predicting All-Cause 30-Day Hospital Readmission? Evidence From Administrative Data. *Value Health* **23**, 1307–1315 (2020).

11. Liu, W. *et al.* Predicting 30-day hospital readmissions using artificial neural networks with medical code embedding. *PLoS One* **15**, e0221606 (2020).

12. Min, X., Yu, B. & Wang, F. Predictive Modeling of the Hospital Readmission Risk from Patients' Claims Data Using Machine Learning: A Case Study on COPD. *Sci. Rep.* **9**, 2362 (2019).

13. Rajkomar, A. *et al.* Scalable and accurate deep learning with electronic health records. *NPJ Digit Med* **1**, 18 (2018).

14. Reddy, B. K. & Delen, D. Predicting hospital readmission for lupus patients: An RNN-LSTM-based deep-learning methodology. *Comput. Biol. Med.* **101**, 199–209 (2018).

15. Wang, H. *et al.* Predicting Hospital Readmission via Cost-Sensitive Deep Learning. *IEEE/ACM Trans. Comput. Biol. Bioinform.* **15**, 1968–1978 (2018).

16. Zhang, D., Yin, C., Zeng, J., Yuan, X. & Zhang, P. Combining structured and unstructured data for predictive models: a deep learning approach. *BMC Med. Inform. Decis. Mak.* **20**, 280 (2020).

17. Ashfaq, A., Sant'Anna, A., Lingman, M. & Nowaczyk, S. Readmission prediction using deep learning on electronic health records. *J. Biomed. Inform.* **97**, 103256 (2019).

18. Golmaei, S. N. & Luo, X. DeepNote-GNN: predicting hospital readmission using clinical notes and patient network. in *Proceedings of the 12th ACM Conference on Bioinformatics, Computational Biology, and Health Informatics* 1–9 (Association for Computing Machinery, 2021).

19. Johnson, A. *et al.* MIMIC-IV (version 1.0). (2021) doi:10.13026/s6n6-xd98.

20. Goldberger, A. L. *et al.* PhysioBank, PhysioToolkit, and PhysioNet: components of a new research resource for complex physiologic signals. *Circulation* **101**, E215–20 (2000).

21. Chen, T. & Guestrin, C. XGBoost: A Scalable Tree Boosting System. in *Proceedings of the 22nd ACM SIGKDD International Conference on Knowledge Discovery and Data Mining* 785–794 (Association for Computing Machinery, 2016).

22. Hamilton, W., Ying, Z. & Leskovec, J. Inductive representation learning on large graphs. *Adv. Neural Inf. Process. Syst.* **30**, (2017).

23. Hochreiter, S. & Schmidhuber, J. Long short-term memory. *Neural Comput.* **9**, 1735–1780 (1997).

24. DeLong, E. R., DeLong, D. M. & Clarke-Pearson, D. L. Comparing the areas under two or more correlated receiver operating characteristic curves: a nonparametric approach. *Biometrics* **44**, 837–845 (1988).

25. Ying, R., Bourgeois, D., You, J., Zitnik, M. & Leskovec, J. GNNExplainer: Generating Explanations for Graph Neural Networks. *Adv. Neural Inf. Process. Syst.* **32**, 9240–9251 (2019).

26. Youden, W. J. Index for rating diagnostic tests. *Cancer* **3**, 32–35 (1950).

27. Schisterman, E. F., Perkins, N. J., Liu, A. & Bondell, H. Optimal cut-point and its corresponding Youden Index to discriminate individuals using pooled blood samples. *Epidemiology* **16**, 73–81 (2005).

28. Schriml, L. M. *et al.* Human Disease Ontology 2018 update: classification, content and workflow expansion. *Nucleic Acids Res.* **47**, D955–D962 (2019).

29. Johnson, A. E. W. *et al.* MIMIC-CXR-JPG, a large publicly available database of labeled chest radiographs. *arXiv [cs.CV]* (2019).

30. Johnson, A. *et al.* MIMIC-CXR-JPG - chest radiographs with structured labels (version 2.0.0). (2019) doi:10.13026/8360-t248.

31. Shuman, D. I., Narang, S. K., Frossard, P., Ortega, A. & Vandergheynst, P. The emerging field of signal processing on graphs: Extending high-dimensional data analysis to networks and other irregular domains. *IEEE Signal Process. Mag.* **30**, 83–98 (2013).

32. Tariq, A. *et al.* Patient-specific COVID-19 resource utilization prediction using fusion AI model. *NPJ Digit Med* **4**, 94 (2021).

33. Huang, G., Liu, Z., Van Der Maaten, L. & Weinberger, K. Q. Densely connected convolutional networks. in *Proceedings of the IEEE conference on computer vision and pattern recognition* 4700–4708 (openaccess.thecvf.com, 2017).

34. He, K., Fan, H., Wu, Y., Xie, S. & Girshick, R. Momentum contrast for unsupervised visual representation learning. in *Proceedings of the IEEE/CVF conference on computer vision and pattern recognition* 9729–9738 (openaccess.thecvf.com, 2020).

35. Irvin, J. *et al.* CheXpert: A Large Chest Radiograph Dataset with Uncertainty Labels and Expert Comparison. *AAAI* **33**, 590–597 (2019).

36. Sowrirajan, H., Yang, J., Ng, A. Y. & Rajpurkar, P. MoCo Pretraining Improves Representation and Transferability of Chest X-ray Models. in *Proceedings of the Fourth Conference on Medical Imaging with Deep Learning* (eds. Heinrich, M. et al.) vol. 143 728–744 (PMLR, 07--09 Jul 2021).

37. Cho, K., van Merrienboer, B., Bahdanau, D. & Bengio, Y. On the Properties of Neural Machine Translation: Encoder-Decoder Approaches. *arXiv [cs.CL]* (2014).

38. Kingma, D. P. & Ba, J. Adam: A Method for Stochastic Optimization. in *3rd International Conference on Learning Representations, ICLR 2015, San Diego, CA, USA, May 7-9, 2015, Conference Track Proceedings* (eds. Bengio, Y. & LeCun, Y.) (2015).

39. Loshchilov, I. & Hutter, F. SGDR: Stochastic Gradient Descent with Warm Restarts. *arXiv [cs.LG]* (2016).

**Supplementary Materials**

# Supplementary Figures

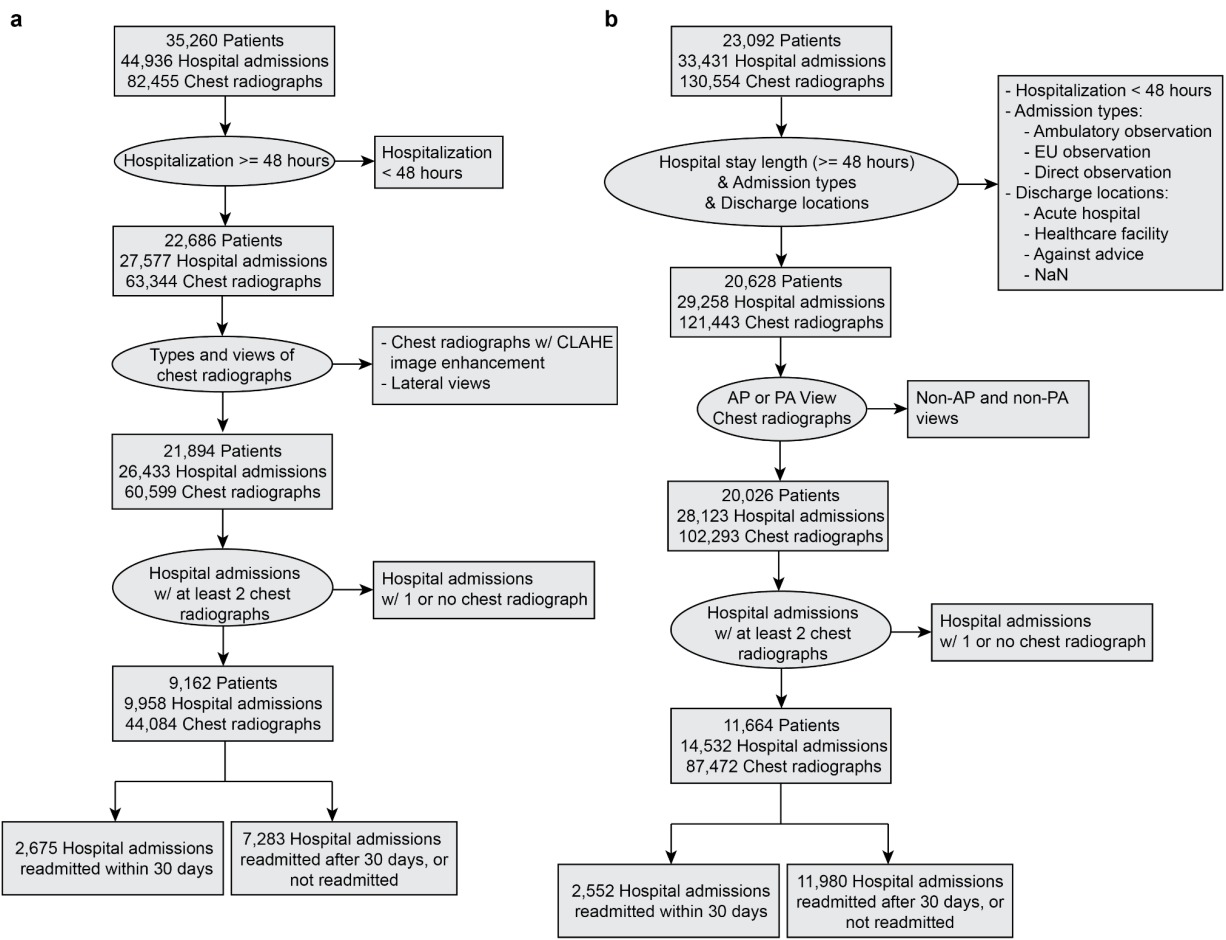

**Supplementary Figure 1.** Inclusion and exclusion criteria for the **(a)** primary dataset and **(b)** MIMIC-IV dataset.

# Supplementary Tables

**Supplementary Table 1.** 30-day all-cause readmission prediction results for baselines and MM-STGNN on disease-specific subsets in the test set of MIMIC-IV. Confidence intervals (CI) were calculated using the DeLong method[1].

| Model | Modality | Diseases of Circulatory System (n = 356) AUROC [95% CI] | Diseases of Immune Mechanism (n = 165) AUROC [95% CI] | Endocrine, Nutritional and Metabolic Diseases (n = 117) AUROC [95% CI] |
|---|---|---|---|---|
| XGBoost | Imaging | 0.663 [0.594-0.732] | 0.716 [0.632-0.799] | 0.581 [0.450-0.713] |
| GNN | Imaging | 0.617 [0.542-0.691] | 0.680 [0.576-0.784] | 0.524 [0.387-0.660] |
| LSTM | Imaging | 0.702 [0.618-0.787] | 0.707 [0.600-0.813] | 0.765 [0.652-0.879] |
| XGBoost | EHR | 0.721 [0.646-0.795] | 0.684 [0.569-0.798] | 0.737 [0.620-0.853] |
| GNN | EHR | 0.766 [0.688-0.844] | 0.731 [0.615-0.847] | 0.718 [0.607-0.829] |
| LSTM | EHR | 0.784 [0.716-0.851] | 0.768 [0.670-0.866] | 0.651 [0.525-0.776] |
| MM-STGNN | EHR+Imaging | 0.821 [0.751-0.890] | 0.763 [0.653-0.872] | 0.789 [0.671-0.908] |

**Supplementary Table 2.** Comparison of single modality STGNN performance using imaging or EHR as node features and MM-STGNN performance. Confidence intervals (CI) are calculated using the DeLong method[1].

| Model | Modality | Primary Dataset AUROC [95% CI] | MIMIC-IV AUROC [95% CI] | Primary Dataset LACE+ Subset AUROC [95% CI] |
|---|---|---|---|---|
| STGNN | Imaging | 0.725 [0.701-0.749] | 0.726 [0.701-0.752] | 0.649 [0.578-0.721] |
| STGNN | EHR | 0.786 [0.763-0.809] | 0.781 [0.757-0.805] | 0.792 [0.734-0.851] |
| MM-STGNN | EHR+Imaging | 0.788 [0.765-0.811] | 0.791 [0.766-0.816] | 0.772 [0.709-0.834] |

**Supplementary Table 3.** Comparison of single modality STGNN performance using imaging or EHR as node features and MM-STGNN performance on disease-specific subsets in the primary dataset. Confidence intervals (CI) are calculated using the DeLong method[1].

| Model | Modality | Heart Disease (n = 843) AUROC [95% CI] | Vascular Disease (n = 825) AUROC [95% CI] | Metabolism Disease (n = 548) AUROC [95% CI] |
|---|---|---|---|---|
| STGNN | Imaging | 0.755 [0.720-0.790] | 0.726 [0.689-0.762] | 0.690 [0.644-0.736] |
| STGNN | EHR | 0.790 [0.757-0.824] | 0.787 [0.753-0.821] | 0.751 [0.707-0.795] |
| MM-STGNN | EHR+Imaging | 0.800 [0.765-0.834] | 0.786 [0.751-0.821] | 0.747 [0.703-0.792] |

**Supplementary Table 4.** Comparison of non-temporal multimodal GNN (MM-GNN) performance and MM-STGNN performance. Confidence intervals (CI) are calculated using the DeLong method[1].

| Model | Temporal Model | Modality | Primary Dataset AUROC [95% CI] | MIMIC-IV AUROC [95% CI] | Primary Dataset LACE+ Subset AUROC [95% CI] |
|---|---|---|---|---|---|
| MM-GNN | No | EHR+Imaging | 0.755 [0.730-0.780] | 0.753 [0.728-0.778] | 0.717 [0.646-0.788] |
| MM-STGNN | Yes | EHR+Imaging | 0.788 [0.765-0.811] | 0.791 [0.766-0.816] | 0.772 [0.709-0.834] |

**Supplementary Table 5.** MM-STGNN performance using different EHR sources for the edges. "-" indicates either the EHR source is not available or the resulting graph is too big to fit into GPU memory. "All" means concatenation of all available EHR sources.

| Model | Modality for Edge | Primary Dataset AUROC [95% CI] | MIMIC-IV AUROC [95% CI] |
|---|---|---|---|
| MM-STGNN | Demo | 0.779 [0.756-0.803] | 0.791 [0.766-0.816] |
| | CPT | 0.788 [0.765-0.811] | - |
| | ICD | 0.771 [0.747-0.795] | - |
| | Lab | - | 0.791 [0.766-0.816] |
| | Medication | 0.780 [0.757-0.803] | 0.782 [0.756-0.807] |
| | All | 0.770 [0.746-0.795] | 0.775 [0.750-0.800] |

**Supplementary Table 6.** Top 50 primary diagnoses and their frequency counts in our primary dataset.

| Primary Diagnoses | Count |
|---|---|
| Heart disease | 4348 |
| Vascular disease | 4219 |
| Disease of metabolism | 2843 |
| Kidney disease | 1431 |
| Cancer | 1321 |
| Sleep disorder | 1184 |
| Uterine cancer | 1022 |
| Mitral valve stenosis | 989 |
| Overnutrition | 986 |
| Aortic valve prolapse | 937 |
| Disease of mental health | 844 |
| Intestinal disease | 744 |
| Blood coagulation disease | 742 |
| Anemia | 671 |
| Disease of cellular proliferation | 609 |
| Aortic valve stenosis | 604 |
| Bacterial infectious disease | 513 |
| Respiratory system disease | 498 |
| Lung disease | 463 |
| Neuropathy | 428 |
| Brain disease | 358 |
| Aortic disease | 358 |
| Genetic disease | 314 |
| Substance-related disorder | 305 |
| Renal tubular transport disease | 284 |
| Immune system disease | 268 |
| Nutrition disease | 267 |
| Bone disease | 262 |
| Nervous system disease | 256 |
| Peripheral nervous system disease | 236 |
| Substance abuse | 232 |
| Muscular disease | 191 |
| Bacterial pneumonia | 186 |
| Cellulitis | 180 |
| Autoimmune disease of the nervous system | 178 |
| Liver disease | 170 |
| Autoimmune disease | 156 |
| Carcinoma | 156 |
| Disease by infectious agent | 151 |
| Thoracic disease | 148 |
| Physical disorder | 147 |
| Mouth disease | 134 |
| Cardiovascular system disease | 126 |
| Urinary tract obstruction | 125 |
| Bicuspid aortic valve disease | 123 |
| Pancreatitis | 118 |
| B-cell lymphoma | 116 |
| Gonadal disease | 110 |
| Parathyroid gland disease | 108 |
| Syndrome | 106 |

# Supplementary References

1. DeLong, E. R., DeLong, D. M. & Clarke-Pearson, D. L. Comparing the areas under two or more correlated receiver operating characteristic curves: a nonparametric approach. *Biometrics* **44**, 837–845 (1988).